# A Hybrid Citation Retrieval Algorithm for Evidence-based Clinical Knowledge Summarization: Combining Concept Extraction, Vector Similarity and Query Expansion for High Precision

Kalpana Raja, PhD[1], Andrew J Sauer, MD[2,3], Ravi P Garg, MSc[1], Melanie R Klerer[1], Siddhartha R Jonnalagadda, PhD[1*]

[1]Division of Health and Biomedical Informatics, Department of Preventive Medicine, Northwestern University Feinberg School of Medicine, Chicago, IL

[2]Bluhm Cardiovascular Institute of Northwestern Medicine, Northwestern University Feinberg School of Medicine, Chicago, IL

[3]Center for Advanced Heart Failure & Heart Transplant, University of Kansas Medical Center, Kansas City, KS

[*]Corresponding author

Dr. Siddhartha Jonnalagadda, PhD

Division of Health and Biomedical Informatics

Department of Preventive Medicine

Northwestern University Feinberg School of Medicine

Chicago, IL

E-mail: sid@northwestern.edu

Telephone: 312-503-2826

**Abstract**

Novel information retrieval methods to identify citations relevant to a clinical topic can overcome the knowledge gap existing between the primary literature (MEDLINE) and online clinical knowledge resources such as UpToDate. Searching the MEDLINE database directly or with query expansion methods returns a large number of citations that are not relevant to the query. The current study presents a citation retrieval system that retrieves citations for evidence-based clinical knowledge summarization. This approach combines query expansion, concept-based screening algorithm, and concept-based vector similarity. We also propose an information extraction framework for automated concept (Population, Intervention, Comparison, and Disease) extraction. We evaluated our proposed system on all topics (as queries) available from UpToDate for two diseases, heart failure (HF) and atrial fibrillation (AFib). The system achieved an overall F-score of 41.2% on HF topics and 42.4% on AFib topics on a gold standard of citations available in UpToDate. This is significantly high when compared to a query-expansion based baseline (F-score of 1.3% on HF and 2.2% on AFib) and a system that uses query expansion with disease hyponyms and journal names, concept-based screening, and term-based vector similarity system (F-score of 37.5% on HF and 39.5% on AFib). Evaluating the system with top K relevant citations, where K is the number of citations in the gold standard achieved a much higher overall F-score of 69.9% on HF topics and 75.1% on AFib topics. In addition, the system retrieved up to 18 new relevant citations per topic when tested on ten HF and six AFib clinical topics. These citations were verified by a cardiologist to provide potentially useful and new information not available in UpToDate. For these sixteen topics, the concept-based screening framework in our system retrieved citations at an F-score of 50.6% on HF and 49.3% on AFib. This is significantly higher than an existing PICO concept-based information retrieval system that achieved an F-score of 6.0% on HF and 5.1% on AFib for the same set of topics. The proposed system differs from the existing approaches by focusing on obtaining a high precision while maintaining the recall to obtain the most relevant citations for evidence-based clinical knowledge summarization. We showed that a hybrid combination of query-expansion, concept-based screening and concept-based vector similarity outperformed other approaches.

# INTRODUCTION

The enormous growth of research in the field of medicine is evident with the ever-increasing number of scientific publications in the last decade. More than 700,000 biomedical primary literature articles were added to MEDLINE in 2014 alone.[1] Search engines such as MEDLINE[1] or enhancements such as HubMed[2] supplement the biomedical literature database and provide instant access to citations. However, not all retrieved citations are relevant to specific needs such as evidence-based clinical knowledge summarization. The existing approaches in the biomedical domain widely adapt Medical Subject Headings (MeSH), term frequency weighing, and sentence-level co-occurrence.[3,4] However, significant challenges remain in presenting the most relevant citations for clinicians at point of care.

A previously published clinical knowledge summarization system[5] (that uses query expansion based on Unified Medical Language System (UMLS) Metathesaurus[6]) achieved a very low precision for certain clinical topics in UpToDate[7], a widely used point of care clinical knowledge resource, when compared to citations already present in UpToDate. This is because a number of domain experts manually review, appraise and synthesize evidence from MEDLINE citations. This is extremely time consuming. On the other hand, most existing automated systems return citations whose abstracts include a few terms from the clinical topic. These approaches do not take into account factors such as: (i) whether the abstract includes information that is relevant to the population group of interest, (ii) whether the abstract is specific to the intervention or comparison of interest, and (iii) whether the relevant information is presented as a conclusion or mentioned as a method or background. Other existing information retrieval approaches use different similarity metrics such as tf-idf[8,9] and present citations that rank high based on the metrics. Such approaches are also not reliable for evidence-based clinical knowledge summarization as long as the ranking algorithm does not consider the above-mentioned factors.

Citation retrieval using PICO[10] framework has already been shown to perform comparatively well when PICO concepts from the query are manually entered in a structured format.[11] However, translating a query to PICO format is challenging[12] and abstracting PICO concepts from the unstructured text of the citation is even more complicated. We overcome the limitations by developing two extraction algorithms that uses syntactic regular expressions and custom-tailored dictionaries for extracting PICO concepts. One of these algorithms extracts population, and the other extracts intervention or comparison, and disease.

The purpose of this study is to develop and evaluate an approach for tailoring citations relevant to evidence-based clinical knowledge summarization. We extend an existing query-expansion based system[5] with disease hyponyms, journal names, concept-based screening algorithm, and concept-based vector space model.[13] The concepts considered in the current study are population, intervention or comparison, and disease. The concept-based screening algorithm filters the citations by considering whether their abstracts prominently include information that is relevant to the population group of interest, and intervention, comparison and disease of interest. The vector space model further measures the similarity between the query and citation based on concepts (as opposed to just terms or words themselves).

## BACKGROUND

**Evidence-based medicine and online clinical knowledge resources**

Evidence-based medicine (EBM) is a systematic approach to integrate the best available research evidence into clinical practice.[14] The citations supporting clinical knowledge are researched, created and continually updated from high quality studies published in biomedical journals to provide the treatment options for individual patients or population groups. Integration, organization and utilization of the huge number of available clinical information are necessary to advance health care. Clinicians are expected to have broad knowledge of alternative treatments and diagnoses for a wide array of diseases in order to recommend the most suitable option for their patients. Online clinical knowledge resources such as UpToDate and National Guideline ClearingHouse[15] are available to meet clinicians' information needs at point of care. UpToDate,[7] one of the most commonly used online clinical knowledge resources, summarizes actionable treatment recommendations for more than 10,500 clinical topics in 22 specialties from over 400,000 MEDLINE citations. It is updated every 4 to 6 months with contributions from more than 6,000 physician authors, editors and peer reviewers.[7,16] The recommendations are from a number of resources, including, but not limited to 450 peer-reviewed journals, proceedings of major national and international scientific meetings, guidelines, and databases such as The Cochrane Library.[16]

**MEDLINE Information Retrieval**

MEDLINE or its interface, PubMed uses query-expansion algorithms[3] to refine a simple query using Boolean logic and MeSH vocabulary. Users are able to filter the results based on core clinical journals, clinical queries, publication date, etc.[17] MeSH[18] is a hierarchical vocabulary of medical concepts from the National Library of Medicine (NLM) used for indexing citations in MEDLINE. Abridged Index Medicus from MEDLINE provides a set of 119 core clinical journals.[19] MEDLINE clinical queries retrieve smaller subset of citations of methodologically sound studies meeting evidence-based standards of adult general medicine.[20] Extensions to MEDLINE such as HubMed[2] searches the query terms within a title or an abstract; Relemed[21] uses sentence-level co-occurrence for multi-word queries; and GoPubMed[22] classifies citations on Gene Ontology terms.

**PICO Framework**

Formulating a well-focused query is the most important step in EBM. Identifying relevant citations with appropriate clinical study is highly challenging without a well-focused query. Clinicians use a specialized framework called PICO for exploring sections of a clinical query that are more applicable to their patients.[10,12,23] PICO stands for Patient/Population – Intervention – Comparison – Outcome. PICO framework is expanded to PICOTT (Patient/Population – Intervention – Comparison – Outcome – Type of question being asked – Type of study design),[11] PECODR (Patient/Population/Problem – Exposure/Intervention – Comparison – Outcome - Duration - Results)[24] and PIBOSO (Population – Intervention – Background – Outcome - Study Design – Other).[25] NLM provides a web-based application for PICO with spelling checker for structuring the clinical query.[26] Schardt et.al[11] utilized the NLM PICO interface to improve the MEDLINE search on a given clinical query. Our approach differs from Schardt et.al[11] and is more advanced. We present two extraction algorithms using constituency-tree based patterns, UMLS Metathesaurus,[6] and KEGG's USP drug classification[27] for automatically extracting population, intervention or comparison, and disease concepts. The extracted concepts are used for retrieving and tailoring the citations and for measuring the similarity between the clinical query and citations using vector space model.

**Vector Space Model**

Vector Space Model (VSM) is a mathematical model often used by information retrieval (IR) systems.[13] The relevance measure with VSM is shown to be effective in the context of IR systems.[28,29] The degree of similarity between the input query q and retrieved citation $c_i$ (represents a given citation in a set of retrieved citations) is calculated using a similarity metric such as cosine (θ). The cosine similarity metric is represented as a ratio of the dot product of two vectors q and $c_i$ to their magnitude.[30]

$$sim\theta(q, c_i) = \frac{\vec{q} \cdot \vec{c_i}}{|\vec{q}| \; |\vec{c_i}|}$$

The term frequency-inverse document frequency (tf-idf) is the most common weighing method used for representing document relevancy in VSM.[8,31] Each term in the vector is assigned a weight to represent its importance in $c_i$ and q within the entire set of retrieved citations. The term weight is determined by two factors: the number of occurrences of term j in $c_i$ (term frequency, $tf_{i,j}$) and the total number of occurrences of term j in the entire collection of citations c (document frequency, $df_j$). With N representing the total number of citations, the inverse document frequency ($idf_j$) of a term j can be defined as:

$$idf_j = log \frac{N}{df_j}$$

The composite weight for a term (tf-idf) in each citation is calculated as the product of term frequency and inverse document frequency. The inverse document frequency (and thus, the overall weight) of a term that occurs rarely is higher than that of a frequently occurring term.

$$w_{i,j} = tf_{i,j} \times idf_j = tf_{i,j} \times log \frac{N}{df_j}$$

Our approach is significantly different from existing citation retrieval approaches that use tf-idf.[32-34] Our vectors are not the terms present in the document, but three independent vectors of population, intervention or comparison, and disease concepts.

## METHODS

### System Framework

Figure 1 provides the overview of the proposed citation retrieval system and consists of three stages: (1) query building with NLM's E-Utilities;[35] (2) screening the citations using information extraction framework for population, intervention or comparison, and disease concepts; and (3) ranking with VSM.

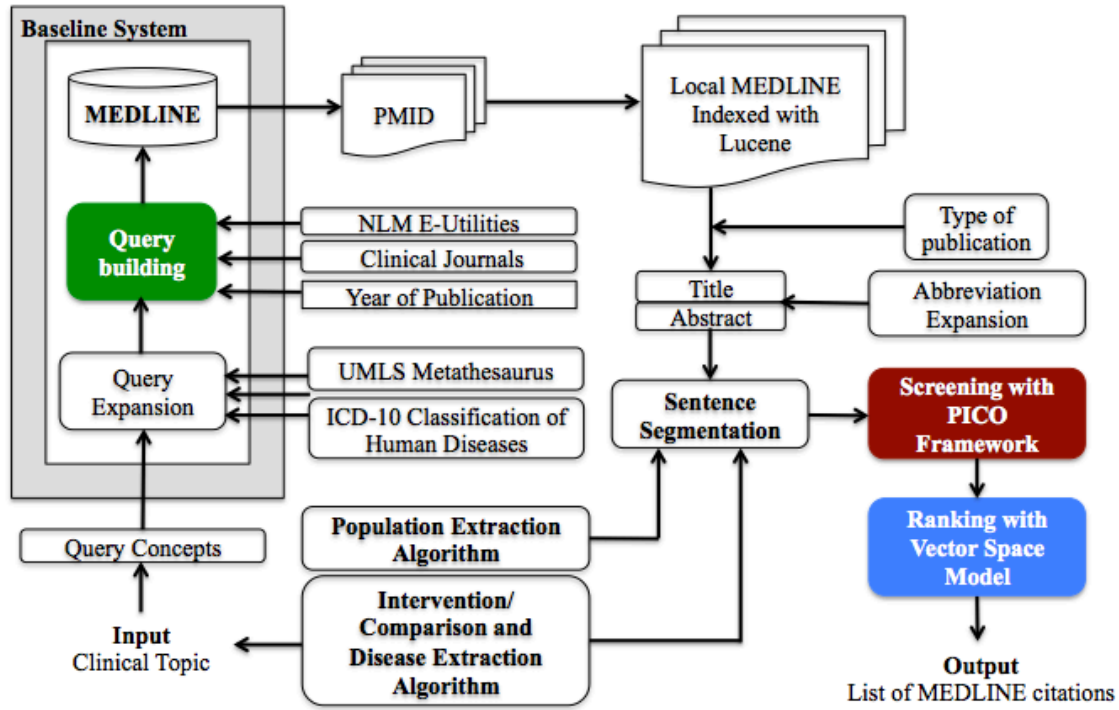

Figure 1. System Architecture

**Query building with NLM's E-Utilities**

The initial retrieval of relevant citations is achieved with the NLM Entrez Programming Utilities (E-Utilities).[35] An automated query-building component is developed based on: (1) the interventions or comparisons, and diseases from the input query, (2) list of high quality clinical journal or proceedings cited in UpToDate, (3) the year of publication and (4) type of publication. The search is executed as a Boolean model.

To automatically modify the input query, we first extracted the intervention or comparison, and disease by mapping to UMLS Metathesaurus[6] using a pipeline of Natural Language Processing (NLP) components combining rule-based and machine learning techniques.[36] The intervention or comparison, and disease concepts are automatically transformed into MeSH terms and added to the query. Hyponyms of each disease are extracted from World Health Organization (WHO) International Classification of Diseases (ICD)-10 classification of human diseases by mapping the text to its parent node.[37] These disease mentions are automatically transformed into MeSH terms and added to the query.

Information from high quality clinical journals is more reliable than from other journals. We revise the query automatically so that it only returns articles published in a set of 200 high quality clinical journals from MEDLINE[1] and McMaster Plus Database[38], 10 journals identified through a journal prioritization algorithm[39] and 642 additional journals and proceedings of major national and international scientific meetings cited by UpToDate. Additionally, based on empirical observation of existing citations in UpToDate, we restricted the articles published before 1974.

Type of publication includes a study with a known study design (systematic review, randomized controlled trial, multiple time series, nonrandomized trial, cohort, case-control, time series, cross-sectional, and case studies),[32] practice guideline or editorial. Articles that do not mention any of these are excluded from the results since they are likely to be non-clinical publications or not peer-reviewed (book chapters, newspaper articles, etc.). Our previous study infers the study design using information present in several sections of a citation, publication type (available from PubMed), MeSH terms, title and abstract.[32] The algorithm infers type of publication first from publication type index and then from MeSH index for the citation. If type of publication is not available in both these indexes, the algorithm screens title and abstract sections for its mention.

**Screening with PICO framework**

Retrieval using E-Utilities is insufficient to identify relevant citations for a specific purpose. It is challenging to design a search query that returns most of the relevant citations without retrieving many irrelevant citations simultaneously. We approach this challenge by developing two extraction algorithms to search for citations matching the population, intervention or comparison, and disease concepts in the title or abstract with the corresponding concepts in the query.

*Preprocessing*

The algorithms for concepts extraction operate at sentence level. An initial preprocessing of abstracts is carried out to segment sentences and replace abbreviations with expanded versions in all the sentences. The use of abbreviations is common in MEDLINE citations, and it is usually defined when it is first used. The primary declaration of an abbreviation mostly appears within a set of parenthesis succeeding its original term (e.g. atrial fibrillation (AFib)). We used the below formula,[40] to obtain max(words), which is the range of words preceding the abbreviation that could contain the expanded form.

$$\text{max(words)} = \min(\text{sum}(ABBR+5), \text{product}(ABBR*2)),$$ where ABBR is the number of characters in the abbreviation.

We extracted the expanded form from max(words) using two constraints: (1) the order of character matching must not change, and (2) the first character of both the expanded form and the abbreviation must match. For example, '<u>A</u>trial <u>F</u>ibrillation' is identified as the expanded form for the abbreviation 'AFib'.

*Extraction of population*

UMLS provides a set of 130 population-related concepts belonging to patient or disabled group semantic type. An additional set of 22 population-related terms was manually identified from MEDLINE citations. We developed a pattern-based algorithm that uses two NLP parsers[41,42] with post-processing rules to identify population within noun or verb phrases. First, sentences from citations are segmented and are parsed with Stanford lexical parser[41] using the PCFG model. The parser generates a constituency parse tree of noun and verb phrases. A sentence may contain more than one noun or verb phrase, and sometimes it is nested within another noun or verb phrase. Then, the generated parse tree is queried using Tregex,[42] a tree query language implemented in Stanford parser (Figure 2). We developed a set of Tregex patterns (Table 1) to extract a sub-tree with population. These patterns are similar to regular expressions and are easy to use. However, the sub-tree matching a Tregex pattern do not always account for population. We use the set of 130 population-related concepts and 22 population-related terms to filter the sub-tree with population.

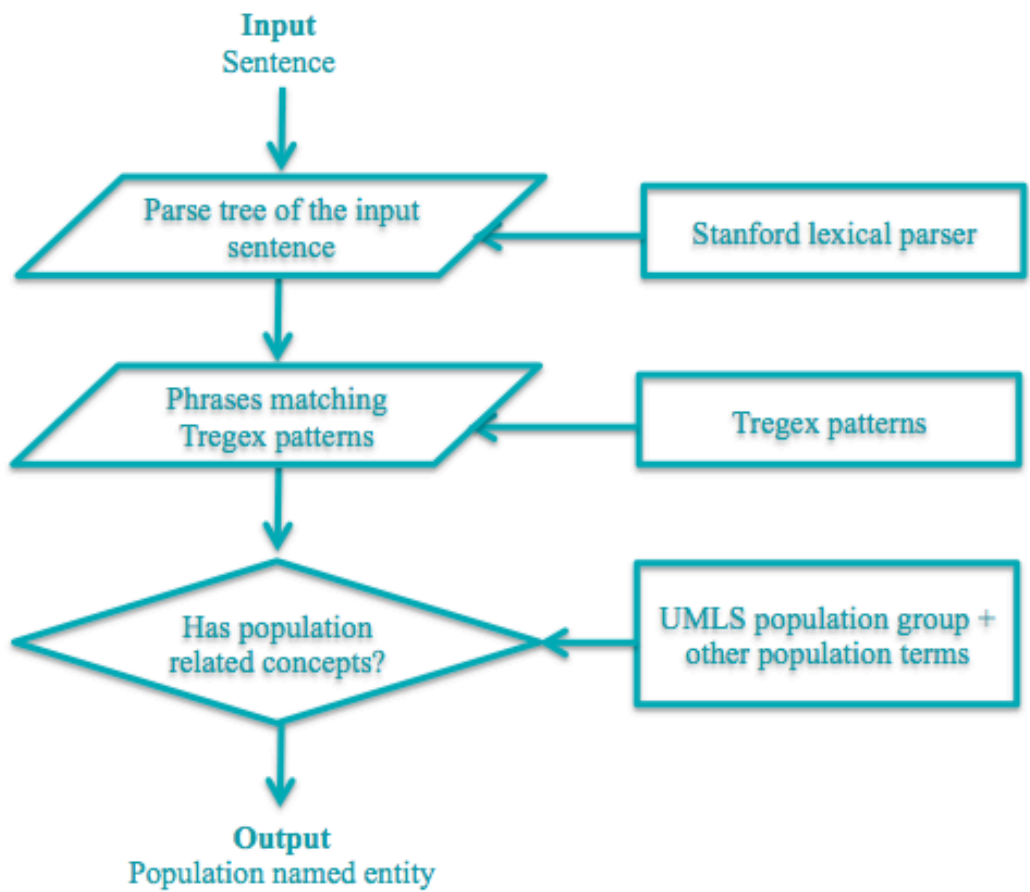

**Figure 2. Population extraction algorithm**

**Table 1. Tregex patterns for population extraction**

| Pattern | Pattern Definition | Post-processing Step | Example Sentence with Output Underlined |
|---|---|---|---|
| 1. NP $ NN | Noun phrase that contains a noun is extracted. | The extracted phrase is screened for the presence of UMLS population concepts or other population terms at the | So far, nebivolol is the only beta-blocker to have been shown effective in <u>*elderly heart failure patients*</u>, regardless of their left ventricular ejection fraction. (PMID: 20307222) |
| | | beginning of the phrase. | |
| 2. NP < VP | Noun phrase along with the succeeding verb phrase is extracted. | The extracted phrase is screened for the presence of UMLS population concepts or other population terms at the beginning of the phrase. | These findings provide further support for the idea that spironolactone may be useful in *patients hospitalized with HF and reduced LVEF*. (PMID: 21146672) |
| 3. NP < SBAR | Noun phrase along with succeeding subordinating conjunction is extracted. | The extracted phrase is screened for the presence of UMLS population concepts or other population terms at the beginning of the phrase. | An improved adverse-effect profile also makes angiotensin II receptor blockers appropriate in *patients who cannot tolerate ACE inhibitors*. (PMID: 14563505) |
| 4. NP < PP | Noun phrase along with succeeding propositional phrase is extracted. | The extracted phrase is screened for the presence of UMLS population concepts or other population terms at the | ACE inhibitors decrease mortality in *patients with heart failure resulting from left ventricular systolic dysfunction*. (PMID: 14727993) |
| | | beginning of the phrase. | |
| 5. @NP | Noun phrase is extracted iteratively. | Each extracted phrase is screened for the presence of UMLS population concepts or other population terms. If present, the algorithm extracts the substring starting from the population concept or term till the end of noun phrase at the beginning of the phrase. | Aldosterone blockade has been shown to be effective in reducing *total mortality as well as hospitalization for heart failure in patients with systolic left ventricular dysfunction (SLVD) due to chronic heart failure and in patients with SLVD post acute myocardial infarction.* (PMID: 15134801) |
| 6<br>6.1 @VP,<br>6.2 @NP | Verb phrase (6.1) that contains at least one noun phrase (6.2) is extracted. | The extracted phrase is screened for the presence of UMLS population concepts or other population terms. If present, the algorithm extracts | HF pharmacotherapies that *have been associated with mortality benefits in elderly patients with left ventricular systolic dysfunction* include ACE inhibitors or ARBs; beta-blockers; aldosterone antagonists; and, in |

| | | the substring starting from the population concept or term till the end of verb phrase. | patients who cannot tolerate ACE inhibitors or ARBs or who are black, a combination of hydralazine and nitrates. (PMID: 19948300) |
|---|---|---|---|
| 7. 7.1 @VP, 7.2 PP < SBAR, 7.3 @NP | Verb phrase (7.1) that contains prepositional phrase and subordinating conjunction (7.2) with nested noun phrase (7.3) is extracted. | The extracted phrase is screened for the presence of UMLS population concepts or other population terms. If present, the algorithm extracts the substring starting from the population concept or term till the end of verb phrase. | Isosorbide dinitrate and hydralazine hydrochloride should be *tried in patients who cannot tolerate ACE inhibitors or who have refractory symptoms*. (PMID: 7933398) |

Patterns 1 to 5 (Table 1) are first developed to extract population concept, a noun phrase. Pattern 5 alone is sufficient to extract the population within a noun phrase. However, the noun phrase patterns sometimes extract incorrect population information since parsing is not 100% accurate. For example, '*elderly patients*' is extracted instead of '*elderly patients with left ventricular systolic dysfunction*' from the sentence '*HF pharmacotherapies that have been associated with mortality benefits in elderly patients* *with left ventricular systolic dysfunction include ACE inhibitors or ARBs; beta-blockers; aldosterone antagonists; and, in patients who cannot tolerate ACE inhibitors or ARBs or who are black, a combination of hydralazine and nitrates*' (PMID: 19948300) with the patterns 1 to 5. To overcome such parsing errors, we developed two verb phrase patterns (6 and 7 from Table 1). The meaning of these patterns and the subsequent post processing steps are in Table 1.

*Extraction of intervention, comparison, and disease*

*Concept Extraction*

Interventions include the use of treatment, diagnostic test, adjunctive therapy, medications (e.g. chemicals and drugs), lifestyle changes (e.g. diet or exercise), and recommendations to the patient or population to use a product or procedure. Comparisons are the main alternative interventions to be considered for a patient or population.[43] Diseases or disorders are the pathological condition of a body part, organ, or system that impairs normal functioning. Interventions, comparisons, and diseases are usually one or more words (e.g. furosemide, ACE inhibitors, heart failure) and their extraction is achieved first by mapping the text to UMLS Metathesaurus (Figure 3(a)). Our approach is a pipeline of NLP components combining rule-based and machine learning techniques.[5] The pipeline consists of four components: tokenization, lexical normalization, UMLS Metathesaurus look-up, and concept screening. The tokenization component splits the query and citation into tokens or words using open NLP suite.[44] The lexical normalization component converts words into canonical form using an efficient in-memory data structure similar to a hash table.[45] UMLS Metathesaurus look-up is performed using a well-known efficient algorithm called Aho-Corasick string matching.[46]

The concept screening component limits UMLS concepts to four semantic groups[47]: disorders, chemicals (includes drugs), procedures, and devices.

*Drug Normalization*

In biomedical literature, drugs are mentioned as a general drug class (e.g. beta-blockers) or as a specific drug (e.g. carvedilol). Such mentions (drug class vs. drug, drug A vs. drug B) are normalized using USP (United States Pharmacopoeia) drugs classification available from Kyoto Encyclopedia of Genes and Genomes (KEGG) Database (Figure 3(b)).[27] We developed a dictionary of drug hierarchy with a depth of three consisting of 2,124 drugs belonging to 49 drug classes using KEGG's USP drugs classification. A snippet of the dictionary is shown in Figure 4. For example, furosemide and bumetanide are recognized as loop diuretics in the leaf node (depth =3), as diuretics in the middle level (depth =2) and as cardiovascular agents in the top level (depth =1).

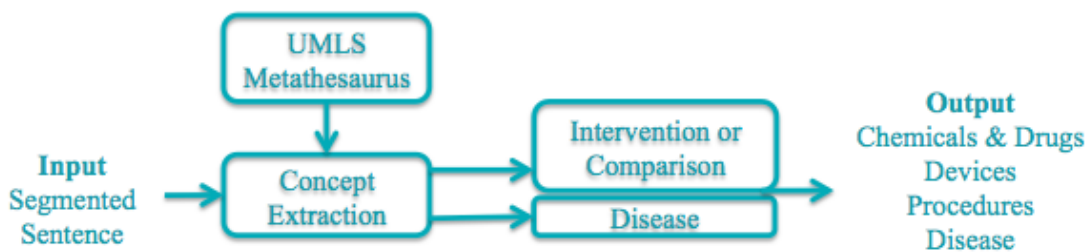

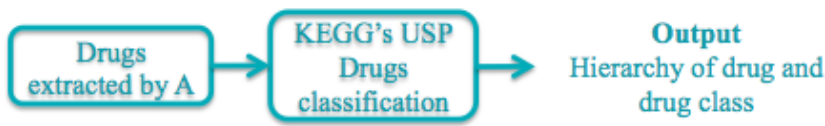

Figure 3. Intervention or Comparison and Disease extraction algorithm

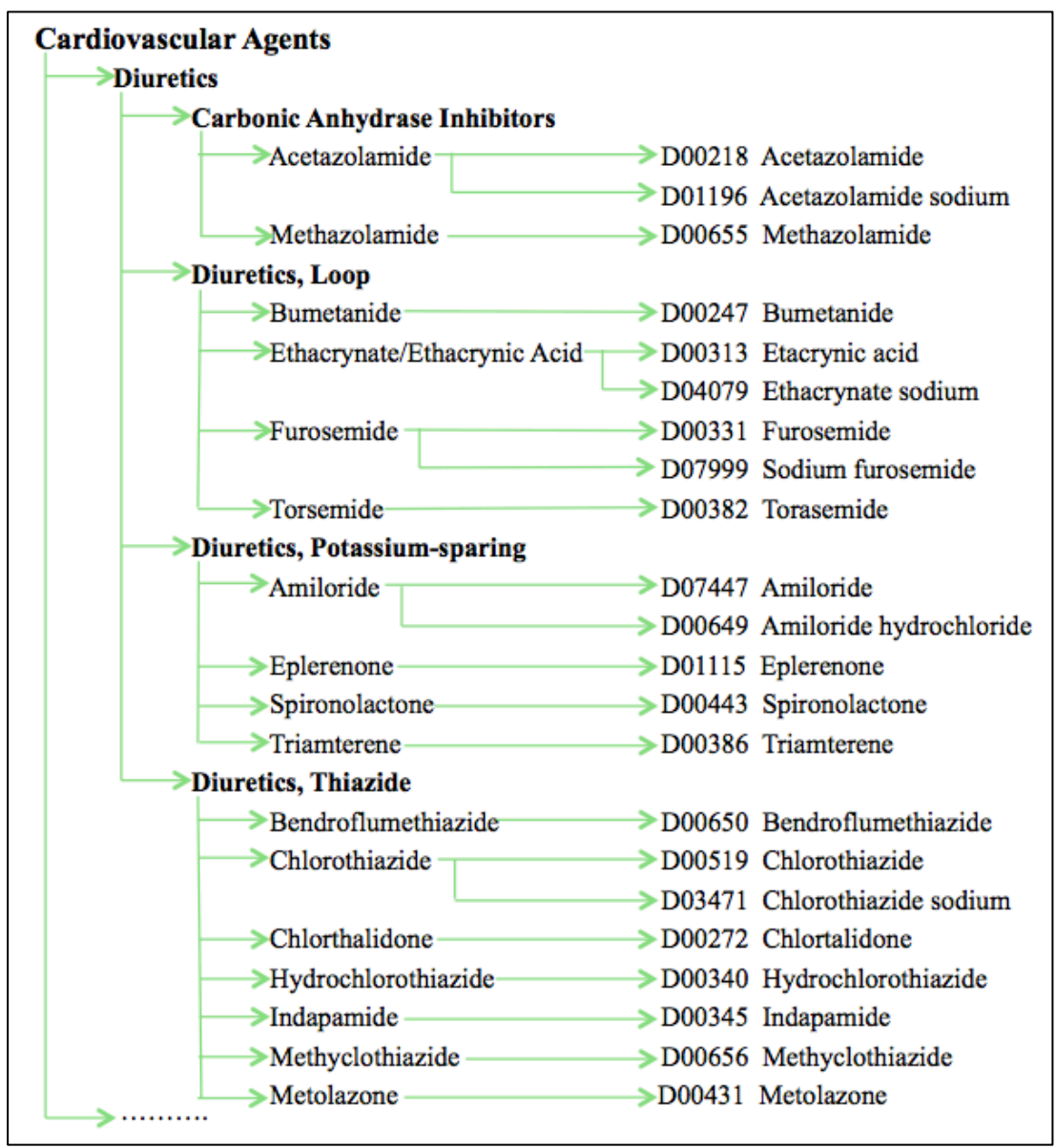

**Figure 4. A snippet showing USP drugs classification**

The orthographical and morphological term variations in chemicals and drugs in the citations and query restrict the effective direct use of the mentioned text. The chemicals and drugs previously extracted with the concept extraction algorithm are processed with a set of rules to attain the uniformity with drug mention in the dictionary (Table 2). Normalization of biomedical concepts using rules has been shown to perform well on MEDLINE citations and other text documents.[48-52] The rules 1 to 4 in Table 2 were derived from our previous study and were shown to provide good performance.[48] The generalizability of these rules is evident from F-score across different datasets: 80.27% on BioCreAtivE-II gene normalization (training set), 83.33% on BioCreAtivE-II gene normalization (test set), and 90.91% on ProNormz human protein kinase gene normalization corpus. In the current study, we used the rules for normalizing drugs. The rules 5 and 6 are specific to drug normalization and derived based on USP drug class terminology (Table 2). For every drug, the hierarchy of drug classes is retrieved and used in subsequent modules for citation screening and ranking.

**Table 2. Rules for dictionary mapping**

| Rule | Example (Original form ➔ Modified form) |
|---|---|
| 1. Case normalization | aldosterone antagonists ➔ Aldosterone antagonists |
| 2. Recognition of Arabic and Roman numerals | Angiotensin receptor blockers ➔ Angiotensin II receptor blockers |
| 3. Recognition of interventions separated by slash, hyphen etc. | Isosorbide dinitrate/Hydralazine ➔ Isosorbide dinitrate, Hydralazine |
| 4. Removal of contents inside parenthesis | Angiotensin converting enzyme (ACE) inhibitors ➔ Angiotensin converting enzyme inhibitors |
| 5. Recognition of equivalent interventions that are in the same drug class | Beta blockers ➔ Beta adrenergic blockers |
| 6. Singular to Plural for drug | Diuretic ➔ Diuretics |

| class | |
|---|---|

*Extraction of outcomes*

Outcomes are the results of a study and most of the clinically relevant outcomes are in relation to a disease (incidence of, remission of, readmission due to, etc.). We have not extracted the outcome concept in this study; however, given that we are extracting disease concepts and limiting to high quality clinical journals and proceedings it might not be necessary to rank or filter the citations based on the outcomes.

*Concept-based screening algorithm*

Using the information extraction algorithms described previously, we extract population, intervention, comparison and disease concepts from the query and the MEDLINE citation. For citations, information from title, abstract and MeSH index are used to extract the required concepts. We use the extraction algorithms to obtain concepts from title and abstract. However, for extracting the concepts from MeSH index, we use the descriptors and their qualifiers.

Each MeSH term (XML format) includes a Descriptor Name (e.g. <DescriptorName …>Heart Failure</DescriptorName>) and an optional Qualifier Name (e.g. <QualifierName …>drug therapy</QualifierName>). The descriptor name is the MeSH term from NLM controlled vocabulary thesaurus and the qualifier name represents the context of the MeSH term in a citation. The qualifier name also mentions whether it is the main topic of the citation (e.g. MajorTopicYN="Y"). The descriptor name and qualifier name together represents the information available in the citation. For example, presence of 'heart failure/drug therapy' (descriptor name/qualifier name) as the main topic indicates that the citation conveys information on drug therapy for heart failure. In the current study, we restricted the qualifier name to the following clinically significant categories: therapy, diagnosis, diagnostic use, drug therapy, mortality, surgery, ultrasonography, prevention and control, rehabilitation, complications, congenital, epidemiology, ethnology, etiology, therapeutic use, pharmacology, adverse effects, contraindications, administration and dosage, agonists, antagonists and inhibitors, and analogs and derivatives. These qualifier names are derived from the entire list of qualifiers (83 in total) from NLM.[53] Certain qualifier names such as blood, blood supply, biosynthesis, chemically synthesis, chemistry, history, metabolism, etc. are not significant.

For each concept extracted from the query using our extraction algorithms, the screening algorithm looks for a match within citation's MeSH index, title and abstract sections. The matching is performed at concept level. A citation is screened in or considered relevant to the query if it satisfies one of the following constraints in the given order:

1. A MeSH term that matches the query has a clinically significant qualifier name and the qualifier name is marked as the main topic.
2. Title contains population, intervention, comparison, and disease relevant to query.
3. The conclusion section of the abstract contains population, intervention, comparison, and disease relevant to query. In structured abstracts, the conclusion section is identified with subheadings (e.g. conclusions, authors' conclusions, reviewers' conclusions, conclusions and relevance, and interpretation). In unstructured abstracts the conclusion section is identified with related phrases (e.g. 'In conclusion', 'We conclude that'). If still not found, the presence of query concepts is checked in the last two sentences.

4. Population, intervention, comparison, and disease relevant to query are present in in the same or subsequent sentences.

**Ranking with Concept-based Vector Space Models**

Many IR systems frequently prioritize the retrieved citations based on journal-related metrics,[54,55] MeSH terms,[56,57] study design,[32] etc. Among the various approaches available, term-based vector similarity is widely adopted and more reliable. In the current study, we introduce concept-based vector similarity instead of term-based vector similarity. Here, the contents of vector are concepts (e.g. population, intervention, comparison and disease) instead of terms. Our approach represents the query or citation as three independent vectors namely population vector, intervention or comparison vector, and disease vector. Extraction of population, intervention or comparison, and disease concepts in the query and citation are achieved with our extraction algorithms discussed above. For population vector, we first performed stemming and removal of stopwords. The similarity between the query and citation is estimated by measuring the similarity between the respective concept vectors.

We constructed two sets of concepts-based vector space models: one for the query and other for the citation. Each set consists of three vectors namely population vector, intervention or comparison vector, and disease vector. The similarity between query and citation is calculated with tf-idf: the similarity in terms of population is estimated as 'Population similarity score', the similarity in terms of intervention or comparison is estimated as 'Intervention similarity score', and the similarity in terms of disease is estimated as 'Disease similarity score'. The overall similarity is estimated as the sum of three similarity scores mentioned above. In addition, we assign weight hyper parameter for each of these similarity scores such that the total weight is always equal to one, w1+w2+w3=1 In the current study intervention or comparison are assigned with a higher weight, i.e. w2=0.4. Disease mentions mostly appear in combination with population (e.g. patients with heart failure) and both are assigned with equal weights, i.e. w1=0.3 and w3=0.3. Various combinations of weight measures for population, intervention or comparison, and disease are tested. The performance of the system remains constant for the tested values (reported in 'Discussion section'). We refer to this similarity measure as *VSM score* between query and citation in terms of population, intervention or comparison, and disease concepts.

$$VSM\ score\ = (population\ similarity\ score * w1)$$
$$+ (intervention\ similarity\ score\ * w2)$$
$$+ (disease\ similarity\ score * w3)$$

**Evaluation Approach**

To evaluate the effectiveness of our system, we performed an experiment that assessed the citations retrieved by the system on all UpToDate topics over two diseases: atrial fibrillation (AFib) and heart failure (HF): 110 topics for AFib and 110 topics for HF. Topics relevant to both diseases (e.g. the management of atrial fibrillation in patients with heart failure) are considered under the disease for which it is more relevant. Topics related to other diseases (e.g. the use of bisphosphonates in postmenopausal women with osteoporosis) or those that are more general (e.g. fish oil and marine omega-3 fatty acids) are not included in the study. The topics were collected after the system development and therefore did not bias the evaluation. For system development, three topics suggested by a cardiologist (AJS) ('Anticoagulation in older adults'; 'Management of thromboembolic risk in patients with atrial fibrillation and chronic kidney disease'; and 'The use of antithrombotic therapy in patients with an acute or prior intracerebral hemorrhage' topics from UpToDate) were used. The initial performance of the system was tested with all sixteen topics from two clinical areas: anticoagulation therapy for atrial fibrillation (6 topics) and treatment for heart failure due to systolic dysfunction (10 topics). As the performance was promising, we evaluated the system on all topics related to AFib (110 topics) and HF (110 topics). Our system does not learn features or use rules that are based on the topics and the topics used for our evaluation are different from the topics used for development. We adapted the standard evaluation metrics of precision, recall, and F-score to measure the system performance. The evaluation metrics is defined as:

$$Precision = \frac{TP}{TP + FP}$$

$$Recall = \frac{TP}{TP + FN}$$

$$F - score = \frac{2 \ (Precision)(Recall)}{Precision + Recall}$$

Here, TP (true positive) represents the number of citations that were correctly retrieved, FP (false positive) represents the number of citations that were incorrectly retrieved, and FN (false negative) represents the number of citations that the system failed to retrieve. We prepared the gold standard for each topic by manually extracting PMIDs (MEDLINE identifier) cited in UpToDate. This gold standard is available for sharing upon obtaining permission from UpToDate.

A second evaluation was performed on citations identified as false positives. We assumed that one or more citations of false positives might be query relevant but are not available in UpToDate (i.e. false negatives of UpToDate). We considered all UpToDate topics from the two clinical areas mentioned above, for performing this evaluation. The citations were rated independently by a cardiologist (AJS), and a pharmacist (KR) from our team according to two attributes: (1) conclusive sentences are relevant to query (sentence (1) in box 1), and (2) citation is not available in any related topics of UpToDate. The inter-annotator accuracy is measured using kappa coefficient.[58] A citation is considered to have new information when it is not cited in UpToDate and contains information that adds new knowledge to UpToDate (sentence (2) in box 1).

| Box 1: Example sentences |
|---|
| Clinical topic: Anticoagulant therapy for atrial fibrillation<br><br>(1) Conclusive sentence: Use of anticoagulation among stroke patients with AF has increased to very high levels overall in GWTG-Stroke over time. (PMID: 21982662)<br><br>(2) New knowledge: The combination of aspirin and clopidogrel is not as effective as oral anticoagulants in patients with atrial fibrillation, whereas the combination of aspirin and clopidogrel is more effective than oral anticoagulants in patients with coronary stents. (PMID: 17635734) |

## RESULTS

**System performance**

Table 3 shows the overall performance of the system on the 220 UpToDate topics in the gold standard that together contains 9,333 citations. We evaluated these in different setting: (1) retrieval using a query expansion based system,[5] (2) retrieval using the query expansion based system, ranked with term-based vector similarity, (3) retrieval using the query expansion based system with disease hyponyms and journal names, ranked with term-based vector similarity, (4) retrieval using the query expansion based system with disease hyponyms and journal names, concept screening, ranked with term-based vector similarity, and (5) retrieval using the query expansion based system with disease hyponyms and journal names, concept screening, ranked with concept-based vector similarity.

**Table 3. Overall system performance**

| System | Heart Failure (%) (110 topics) | | | Atrial Fibrillation (%) (110 topics) | | |
|---|---|---|---|---|---|---|
| | P | R | F | P | R | F |
| A. Query expansion | 0.1 | 21.4 | 0.2 | 0.1 | 21.5 | 0.3 |
| B. Query expansion and ranked with term-based vector similarity | 0.6 | 18.0 | 1.3 | 1.2 | 20.1 | 2.2 |
| C. Query expansion with disease hyponyms and journal names, and ranked with term-based vector similarity | 0.2 | 78.9 | 0.3 | 0.2 | 85.1 | 0.4 |
| D. Query expansion with disease hyponyms and journal names, concept- | 26.8 | 62.3 | 37.5 | 28.0 | 66.7 | 39.5 |

| based screening and ranked with term-based vector similarity | | | | | | |
| --- | --- | --- | --- | --- | --- | --- |
| E. Query expansion with disease hyponyms and journal names, concept-based screening and ranked with concept-based vector similarity | 29.1 | 70.4 | 41.2 | 29.3 | 76.8 | 42.4 |

The F-score of 41.2% on HF topics and 42.4% on AFib topics achieved by the final system (System E in Table 3) is significantly higher when compared to the query expansion based system (System A or baseline) that achieved F-score of 0.2% on HF topics and 0.3% on AFib topics, baseline ranked with term-based vector similarity (System B) that achieved an F-score of 1.3% on HF topics and 2.2% on AFib topics, and baseline with disease hyponyms and journal names, ranked with term-based vector similarity (System C) that achieved an F-score of 0.3% on HF topics and 0.4% on AFib topics. The concept-based vector space model (System E) shows a better F-score on both HF topics (41.2% vs. 37.5%) and AFib topics (42.4% vs. 39.5%) when compared to term-based vector space model (System D). The query expansion algorithm (System B) reported a low recall of 18.0% on HF topics and 20.1% on AFib topics even with term-based vector similarity ranking. Our attempt to improve the recall by expanding the input query with disease hyponyms based on ICD-10 classification (as described in Query Expansion with NLM E-Utilities section) significantly increased the recall to 78.9% on HF topics and 85.1% on AFib topics (System C). For example, an input related to intracerebral hemorrhage achieved recall of 5.7% with the query expansion algorithm (System A). Adding of other hemorrhage subtypes, such as cerebral hemorrhage, intracranial hemorrhage and subarachnoid hemorrhage, in System C increased its recall to 84.3%. The hybrid approach of combining concept-based screening with concept-based vector space models (System E) improved the F-score significantly to 41.2% on HF topics and 42.4% for AFib topics by removing many irrelevant citations (System E). System E reports a decrease in recall (8.5% (78.9 – 70.4) on HF topics and 8.3% (85.1 – 76.8) on AFib topics) with a remarkable increase in precision (28.9% (29.1 – 0.2) on HF topics and 29.1% (29.3 – 0.2) on AFib topics) when compared to System C. The decrease in recall in the attempt to increase precision is mostly due to citations without abstracts.

Across both topics, the automated retrieval system presents 74.1% (6917 citations out of 9333 citations) citations available in UpToDate on various clinical topics. Furthermore, the number of citations retrieved and ranked by the system is reasonable when compared to the number of citations in UpToDate for the same topic.

**Manual Analysis of False Positives**

Manual annotation of false positives in the system generated output on the 16 topics from two clinical areas (described in Methods section) identified up to a maximum of 18 citations per topic with potentially useful information that is not available in UpToDate. This suggests a knowledge gap existing between UpToDate and primary literature. These citations are false negatives of UpToDate rather than false positives of our system. Additionally, our analysis revealed that 90 false positive citations are cited in other related UpToDate topics. These citations are true positives rather than false positives of the system. Recalculating F-score without considering the above mentioned categories as false positives show actual and improved performance of the system (Table 4).

**Table 4. Annotation results on FP citations**

| UpToDate Topic | Precision (%) | Recall (%) | F-score (%) | F-score (after annotation) (%) |
|---|---|---|---|---|
| **Atrial Fibrillation** | | | | |
| Anticoagulation in older adults | 62.4 | 86.3 | 72.4 | 75.9 |
| Management of thromboembolic risk in patients with atrial fibrillation and chronic kidney disease | 93.8 | 80.4 | 86.5 | 93.8 |
| The use of antithrombotic therapy in patients with an acute or prior intracerebral hemorrhage | 64.8 | 84.3 | 73.3 | 76.1 |
| Atrial fibrillation: Anticoagulant therapy to prevent embolization | 47.0 | 82.9 | 60.0 | 68.1 |
| Antiarrhythmic drugs to maintain sinus rhythm in patients with atrial fibrillation: Recommendations | 20.5 | 66.7 | 31.3 | 38.2 |
| Antiarrhythmic drugs to maintain sinus rhythm in patients with atrial | 35.7 | 78.2 | 49.0 | 57.0 |
| fibrillation: Clinical trials | | | | |
| **Heart Failure** | | | | |
| Overview of the therapy of heart failure due to systolic dysfunction | 15.8 | 73.4 | 26.0 | 32.4 |
| Rationale for and clinical trials of beta blockers in heart failure due to systolic dysfunction | 58.3 | 86.7 | 69.7 | 81.3 |
| Inotropic agents in heart failure due to systolic dysfunction | 82.1 | 92.0 | 86.8 | 92.0 |
| Use of beta blockers in heart failure due to systolic dysfunction | 25.2 | 87.9 | 39.2 | 51.3 |
| Hydralazine plus nitrate therapy in patients with heart failure due to systolic dysfunction | 46.7 | 73.7 | 57.1 | 63.6 |
| ACE inhibitors in heart failure due to systolic dysfunction: Therapeutic use | 60.6 | 89.6 | 72.3 | 80.0 |
| Use of digoxin in heart failure due to systolic dysfunction | 68.0 | 82.9 | 74.7 | 82.9 |
| Angiotensin II receptor blocker and neprilysin inhibitor therapy in heart failure due to systolic dysfunction | 25.6 | 85.2 | 39.3 | 45.0 |
| Use of aldosterone antagonists in | 75.8 | 69.4 | 72.5 | 84.7 |
| systolic heart failure | | | | |
| Evaluation and management of asymptomatic left ventricular systolic dysfunction | 71.4 | 47.6 | 57.1 | 78.4 |

Table 5 lists citations identified to include potentially new knowledge for UpToDate. The number of citations with new knowledge range from 2 to 6 for topics related to AFib (median is 3) and 1 to 18 for topics related to HF (median is 3). Three topics namely, 'the use of antithrombotic therapy in patients with an acute or prior intracerebral hemorrhage', 'management of thromboembolic risk in patients with atrial fibrillation and chronic kidney disease' and 'inotropic agents in heart failure due to systolic dysfunction' retrieved no citations with new knowledge. Overall, both annotators agreed 81 citations as providing new knowledge to UpToDate (κ=0.94, 95% confidence interval=0.89 to 0.99).

Table 6 lists citations that provide additional evidence for information that is already in UpToDate. The system retrieved 5 to 7 citations for topics related to atrial fibrillation (median is 6) and 1 to 16 citations for topics related to heart failure (median is 3). Citations with additional evidence are not retrieved for three topics: 'management of thromboembolic risk in patients with atrial fibrillation and chronic kidney disease', 'use of mineralocorticoid receptor antagonists in systolic heart failure', and 'evaluation and management of asymptomatic left ventricular systolic dysfunction'.

**Table 5. Citations with new knowledge**

| UpToDate Topic | New citation (PMID) |
|---|---|
| Anticoagulation in older adults | 22880717; 16844204; 15047034; 24837794 |
| Atrial fibrillation: Anticoagulant therapy to prevent embolization | 17636831; 2407959 |
| Antiarrhythmic drugs to maintain sinus rhythm in patients with atrial fibrillation: Recommendations | 15607398; 19029470; 12010934; 16569550; 8526696 |
| Antiarrhythmic drugs to maintain sinus rhythm in patients with atrial fibrillation: Clinical trials | 14739742; 12010934; 15607398; 19029470; 16569550; 8526696 |
| Overview of the therapy of heart failure due to systolic dysfunction | 12446064; 22300776; 15451149; 10781760; 19643361; 16188524; 18652942; 17239677; 19064026; 19931364; 12585952; 20413029 |
| Rationale for and clinical trials of beta blockers in heart failure due to systolic dysfunction | 15632878; 21035578; 9886708; 15846279; 15459606; 17126654; 15894978; 10551702; 15144943; 15144944; 25399276; 23315907; 19717851 |
| Use of beta blockers in heart failure due to systolic dysfunction | 15632878; 14583895; 17996820; 14760332; |
| | 15144943; 15846279; 10908096; 9886708; 17383287; 15894978; 17126654; 18506054; 15459606; 10551702; 15144944; 25399276; 19717851; 23315907 |
| Hydralazine plus nitrate therapy in patients with heart failure due to systolic dysfunction | 8644661; 6848228 |
| ACE inhibitors in heart failure due to systolic dysfunction: Therapeutic use | 2839020; 9313596; 16520261; 2132302; 24464788; 10502210; 9547444; 6313787; 8869864; 18506054 |
| Use of digoxin in heart failure due to systolic dysfunction | 8376681; 18506054; 19061695; 2537562 |
| Angiotensin II receptor blocker and neprilysin inhibitor therapy in heart failure due to systolic dysfunction | 23219304 |
| Use of mineralocorticoid receptor antagonists in systolic heart failure | 22137068 |
| Evaluation and management of asymptomatic left ventricular systolic dysfunction | 16188524; 19064024 |

**Table 6. Citations with additional evidence for existing knowledge**

| UpToDate Topic | Additional Evidence |
|---|---|
| Anticoagulation in older adults | 10978038; 21621470; 24655744; 23237139; 24733535; 10753981; 24657899 |
| The use of antithrombotic therapy in patients with an acute or prior intracerebral hemorrhage | 4010961; 20733299; 24525481; 20167915; 17290088; 8418549; 22104448; 21748282 |
| Atrial fibrillation: Anticoagulant therapy to prevent embolization | 17158523; 10753981; 10978038; 11601840; 10323820 |
| Antiarrhythmic drugs to maintain sinus rhythm in patients with atrial fibrillation: Recommendations | 18394447; 22032709; 12914883; 24887617; 24728270; 12093058; 21126785 |
| Antiarrhythmic drugs to maintain sinus rhythm in patients with atrial fibrillation: Clinical trials | 22032709; 18394447; 12914883; 24887617; 12093058; 15518618; 21126785 |
| Overview of the therapy of heart failure due to systolic dysfunction | 1687118; 22137068; 16169325; 19026308; 9330125; 9886708; 9886706; 10502210; 15144935; 9207617; 22336795; 25399276; 9412542; 19064024; 12177661; 15459606 |
| Rationale for and clinical trials of beta blockers in heart failure due to systolic dysfunction | 10689267; 15781028; 19026308; 16169325; 9129883; 9743509; 15144941; 14691423; 7906702; 12531415; 9207617; 15144935; 15144936 |
| Inotropic agents in heart failure due to | 2523634; 6823859; 19064024 |
| systolic dysfunction | |
| Use of beta blockers in heart failure due to systolic dysfunction | 11583862; 15781028; 10689267; 9129883; 14691423; 12531415; 9743509; 7906702; 19026308; 15144941; 9207617; 16169325; 15144935; 15144936 |
| Hydralazine plus nitrate therapy in patients with heart failure due to systolic dysfunction | 2981463 |
| ACE inhibitors in heart failure due to systolic dysfunction: Therapeutic use | 8376687; 9294790; 8500237; 11074220; 2996575 |
| Use of digoxin in heart failure due to systolic dysfunction | 19064024; 9886709 |
| Angiotensin II receptor blocker and neprilysin inhibitor therapy in heart failure due to systolic dysfunction | 11113722; 23095984 |

**PICO framework for citation retrieval**

In a previous study, PICO framework has been shown to improve the performance of citation retrieval.[11] The study utilized NLM PICO interface that takes the four PICO concepts (population, intervention, comparison and outcome (optional)) as input. The interface further allows the user to select type of publication: clinical trial, meta-analysis, randomized controlled trial, review, and practice guideline. We evaluated our system that does concept-based screening (System D from Table 3) and NLM PICO interface with population, intervention and comparison concepts. Table 7 shows the performance of our system and NLM PICO interface under these settings: our system (Setting A in Table 7, same as System D from Table 3), our system using population, intervention and comparison concepts manually extracted from the topic title (Setting B in Table 7), NLM PICO interface using concepts automatically extracted from the title using our algorithm (Setting C in Table 7), and NLM PICO interface using manually extracted concepts (Setting D in Table 7). We used sixteen topics from two clinical areas (described in 'manual analysis of false positives' subsection). Setting A achieved an overall F-score of 50.6% on HF topics and 49.3% on AFib topics. Setting B achieved an overall F-score of 52.8% on HF topics and 52.1% on AFib topics. This shows that performance of automated extraction (Setting A) is relatively comparable to performance of manually extracted concepts (Setting B). For NLM PICO interface, the performance of automated extraction (Setting C) vs. performance of manually extracted concepts is also almost the same (Table 7). This confirms that our automated extraction performs equally well as that of using manually extracted concepts.

**Table 7. Performance of our system vs. NLM PICO interface utilized by Schardt et.al[11]**

| Setting | Heart Failure (%) (10 topics) | | | Atrial Fibrillation (%) (6 topics) | | |
|---|---|---|---|---|---|---|
| | P | R | F | P | R | F |
| A. Automated concept extraction, Query expansion with disease hyponyms and journal names, concept-based screening, and term-based vector similarity | 38.2 | 74.7 | 50.6 | 40.1 | 63.9 | 49.3 |
| B. Manually extracted concept, Query expansion with disease hyponyms and journal names, concept-based screening, and term-based vector similarity | 39.5 | 79.7 | 52.8 | 43.1 | 65.9 | 52.1 |
| C. Automated concept extraction, NLM PICO interface | 3.7 | 16.2 | 6.0 | 3.0 | 16.6 | 5.1 |
| D. Manually extracted concept, NLM PICO interface | 3.8 | 17.5 | 6.2 | 3.4 | 19.4 | 5.8 |

The overall performance of our system is significantly high when compared to the overall performance of NLM PICO interface with automated extraction (Setting A vs. Setting C) and manual extraction of concepts (Setting B vs. Setting D) (Table 7). The high performance of our system is due to the following facts:

(1) Our concept screening algorithm checks whether the population mentioned in the citation is relevant to the query. The algorithm also checks whether the citation is specific to intervention or comparison of interest. It also checks whether the relevant information is from conclusive sentences. When these factors are not considered, the performance of existing retrieval approaches are found to be poor (as discussed in introduction section). For example, an overall F-score of 6.0% on HF topics and 5.1% on AFib topics achieved by NLM PICO interface (Setting C from Table 7).

(2) Our system searches for citations belonging to eleven categories of type of publication: systematic review, randomized controlled trial, multiple time series, nonrandomized trial, cohort, case-control, time series, cross-sectional, case studies, practice guideline, and editorial. NLM PICO interface searches for citations belonging to only five categories of type of publication: clinical trial, meta-analysis, randomized controlled trial, review and practice guideline. Furthermore, our system includes other parameters i.e. journals, disease hyponyms and year of publication (described in methods section). Accordingly, the recall of 74.7% on HF topics and 63.9% on AFib topics (Setting A from Table 7) achieved by our system is significantly higher than the recall of 16.2% on HF topics and 16.6% on AFib topics (Setting C from Table 7) achieved by NLM PICO interface.

**Further analysis of extraction algorithms**

*Performance of population extraction algorithm*

For preliminary analysis, the algorithm was evaluated on a gold standard of 4,824 sentences from 18 UpToDate documents and a second gold standard of 714 sentences from MEDLINE citations. UpToDate gold standard includes population information related to coronary artery disease, hypertension, depression, heart failure, diabetes mellitus, and prostate cancer. MEDLINE gold standard includes population information related to congestive HF and AFib. The algorithm achieved 91% precision, 97% recall and 94% F-score on UpToDate gold standard.[59] The algorithm achieved 90% precision, 83% recall and 87% F-score on MEDLINE gold standard. The algorithm performs well for thousands of sentences talking about multiple disease conditions from two different datasets.

*Performance of intervention extraction algorithm*

The algorithm for extracting intervention, comparison and disease consists of two components – concept extraction and drug normalization. The concept extraction algorithm (MedTagger) is adopted from our previous study.[5] Overall, the precision, recall, and F-measure of the tool are 80%, 57%, and 67% respectively for strict evaluation and 94%, 77%, and 84% for relaxed evaluation[60] on the CLEF 2013 shared task. The accuracy of MedTagger on a corpus depends on the contents and accuracy of UMLS Metathesaurus, and the accuracy of lexical normalization resource. These resources are developed by NLM and have been extensively used in several NLP systems. While the system has not been intrinsically evaluated for biomedical abstracts, MedTagger has been used as a component of several literature-mining pipelines.[5,59,61]

For drug normalization algorithm, we performed an intrinsic analysis using a gold standard obtained from 852 MEDLINE citations related to HF and AFib that are cited by the sixteen UpToDate topics used in Table 4. The algorithm achieved a precision of 86.8%, recall of 84.7% and F-score of 85.7%.

**DISCUSSION**

In this study we assessed the feasibility of automatically retrieving relevant citations for evidence-based clinical knowledge summarization. The system retrieved 6917 citations out of 9333 citations from all 220 topics. Error analysis on citations that were not retrieved by the system (false negatives) showed that many of these citations do not have abstracts (e.g. PMID: 22052525, 23741058). Therefore, processing citations at abstract level is unsuccessful in spite of their retrieval by query expansion based baseline system. This resulted in a decrease of system recall by about 8% when compared to query expansion based baseline with disease hyponyms and journal names (Table 3). Such citations are mostly from clinical guidelines and their retrieval needs to be addressed.

Our system performance is promising in terms of retrieving relevant citations that are not available in UpToDate. The system retrieved several new citations with new clinical information and several citations that were cited in related articles. The new citations include new knowledge for UpToDate as agreed between the annotators. The capability of enoximone in improving ventricular function when added to digitalis-diuretics therapy (PMID: 6823859) and prolonging life of many hospitalized patients with heart failure with a specific beta-blocker namely carvedilol (PMID: 15144944) etc. are a few example citations with new knowledge.

Certain citations are retrieved for multiple topics. For example, PMIDs 15632878, 17383287, 10809026, 14760332, 9886708, 15846279, 15459606, 17126654, 15894978, 10551702, 15144943, 15144944, 25399276, 23315907, 19717851 are retrieved for two HF topics: 'use of beta-blockers in heart failure due to systolic dysfunction' and 'rationale for and clinical trials of beta blockers in heart failure due to systolic dysfunction'. One possible reason is that certain concepts of both topics (beta-blockers, heart failure and systolic dysfunction) are the same. Additionally, the system retrieved

PMIDs 14583895, 17996820, 10908096, and 18506054 for 'use of beta-blockers in heart failure due to systolic dysfunction' and PMID 21035578 for 'rationale for and clinical trials of beta blockers in heart failure due to systolic dysfunction'.

**Additional experiments on system performance**

We evaluated the system using precision, recall and F-score. Precision is the fraction of retrieved documents that are relevant and recall is the fraction of relevant documents that are retrieved. The harmonic mean of precision and recall is reported as a single measure called F-measure. Precision and recall are the most frequent and basic measures for information retrieval effectiveness for unranked retrieval.[31] We adopted this approach for evaluating our system because the gold standard is not a ranked list. In addition, we also evaluated the system with two more standard information retrieval evaluation approaches, precision at K and precision-recall curve.[31] System performance on Top K citations where K is the number of citations available for that UpToDate topic reports a significantly higher F-scores of 69.9% for HF topics and 75.1% for AFib topics (Table 8). In this analysis, the value of K is not fixed because the number of citations per topic in the gold standard varies between 1 and 155. Nevertheless, this shows that our system can achieve a much higher precision, recall and F-score when the user supplies the system with the number of citations that need to be retrieved.

**Table 8. Performance of the system on Top K citations**

|  | Precision (%) | Recall (%) | F-score (%) |
|---|---|---|---|
| Heart Failure (110 topics) | 64.4 | 76.4 | 69.9 |
| Atrial Fibrillation (110 topics) | 70.5 | 80.3 | 75.1 |

For precision-recall curve, we estimated the overall performance of the system on HF topics (110 in total) and AFib topics (110 in total) at various numbers of citations. For each number of citations retrieved (subset), precision and recall is calculated. The system performance for both HF topics and AFib topics shows a decrease in precision with an increase in recall (Table 9). The corresponding precision and recall of each subset of retrieved citations is plotted with precision in y-axis and recall in x-axis. Figure 5 shows the precision-recall curve for HF topics and AFib topics.

**Table 9. Precision vs. Recall**

| Number of Citations (%) | HF Topics (110 in total) | | AFib Topics (110 in total) | |
|---|---|---|---|---|
| | Precision (%) | Recall (%) | Precision (%) | Recall (%) |
| Top 10 | 70.4 | 17.3 | 73.9 | 18.9 |
| Top 20 | 65.4 | 30.9 | 68.0 | 34.2 |
| Top 30 | 58.0 | 40.6 | 63.6 | 47.7 |
| Top 40 | 51.7 | 47.9 | 57.0 | 56.9 |
| Top 50 | 46.6 | 53.6 | 50.9 | 63.4 |
| Top 60 | 42.2 | 58.1 | 45.2 | 67.3 |
| Top 70 | 38.8 | 62.2 | 40.2 | 69.8 |
| Top 80 | 35.7 | 65.3 | 36.2 | 71.8 |
| Top 90 | 32.7 | 67.3 | 32.8 | 73.2 |
| All | 29.1 | 70.4 | 29.3 | 76.8 |

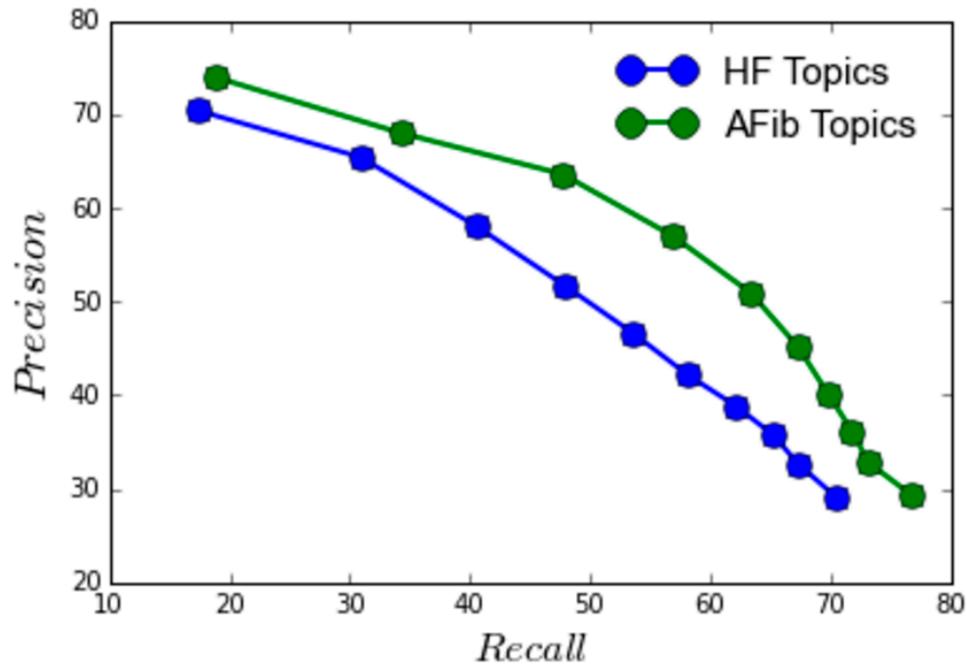

**Figure 5. Precision-Recall curve**

**Salient features of the system**

Approaches that do not consider the three factors (discussed in introduction section) tend to give poor performance on retrieving citations for evidence-based clinical knowledge summarization (e.g. NLM PICO interface (Table 7)). Here, we would like to correlate these factors and our targeted solutions.

Factor 1: MEDLINE citation should include information relevant to the population group of interest.

The screening algorithm checks whether the population information from MEDLINE citation and the query are matching. For achieving this, the population concept from the citation (i.e. title and abstract) and query is extracted with a set of syntactic patterns (described in 'Extraction of population' subsection). The similarity measure is performed at concept level.

Factor 2: The citation should be specific to the intervention, comparison and disease of interest.

The screening algorithm checks whether the intervention, comparison and disease concepts present in the citation are relevant to the query. For achieving this, the required concepts are extracted from MeSH index, title and abstract of the citation. We proposed an algorithm for extracting the concepts from MeSH index (described in 'Screening algorithm' subsection). Extraction of concepts from title and abstract is achieved with our concept extraction algorithm (described in 'Extraction of intervention, comparison and disease' subsection).

Factor 3: The query concepts should be present in the conclusion section of the abstract rather than in background or methods section.

The screening algorithm checks for the presence of query concepts (i.e. population, intervention, comparison and disease) in the conclusion section of a structured abstract. In unstructured abstracts, the conclusion section is identified with related phrases (e.g. 'In conclusion', 'We conclude that'). Otherwise, the presence of query concepts is checked in the last two sentences. Furthermore, the algorithm also checks for the presence of population, intervention, comparison and disease related query concepts in the same or subsequent sentences of the abstract.

**Rule-based drug normalization vs. Fuzzy matching**

In the current study, we used a set of rules to normalize the drug mentions for mapping to the dictionary. Fuzzy matching can also be used for the same task. A challenge with fuzzy matching is that deciding the threshold can be ad hoc; so, we employed intuitive rules instead of an ad hoc score as a dependency. However, this algorithm of ours can be extended for fuzzy matching. We have implemented a string similarity metric to find the similarity between the normalized drug mention in the citation and the normalized dictionary mention in our code. Our approach can be seen as a special case of fuzzy matching where our threshold is 1.0. Our initial experiments showed no improvement in the accuracy if the threshold is decreased to less than 1.0.

**Impact of weight in concept-based vector similarity**

We conducted an analysis on the weight hyper parameter (w1, w2 and w3) by assigning values to w1, w2 and w3 such that their sum is always one, i.e. w1+w2+w3=1 (Table 9). The overall performance of the system remains constant (F-score) for both HF (110 topics) and AFib (110 topics) when the weights are close to each other. However, the overall performance decreased when we considered only one concept (weights of the other two concepts being 0.0). In the current study, we assigned the weights with the following values: w1=w3=0.3 and w2=0.4. If the weights for the concept vectors were all 0.0 and term-based vectors would have been used instead of concept-based vectors, the F-scores for HF and AFib would have reduced from 41.2% and 42.4% to 37.5% and 39.5% respectively (Table 3, System D).

**Table 10. Experiment on weight hyper parameter**

| Weights for individual vectors | | | Heart Failure (110 topics) (F-score) | Atrial Fibrillation (110 topics) (F-score) |
|---|---|---|---|---|
| Population vector (w1) | Intervention vector (w2) | Disease vector (w3) | | |
| 0.3 | 0.4 | 0.3 | 41.2 | 42.4 |
| 0.4 | 0.3 | 0.3 | 41.2 | 42.4 |
| 0.3 | 0.3 | 0.4 | 41.2 | 42.4 |
| 0.33 | 0.33 | 0.33 | 41.2 | 42.4 |
| 1.0 | 0.0 | 0.0 | 39.6 | 41.8 |
| 0.0 | 1.0 | 0.0 | 40.3 | 41.9 |
| 0.0 | 0.0 | 1.0 | 38.3 | 41.0 |

**Study Limitations and future studies**

This study has the following limitations. First, the system evaluation included only two clinical topics, thus limiting the generalizability of our findings. However, the accuracy of the system is consistent across and within the topics. This is evident from a similar overall F-score of 41.2% (95% confidence interval of 31.9% to 43.1%) for HF and 42.4% (95% confidence interval of 32.0% to 43.2%) for AFib. The preliminary results provide useful insights into the feasibility of the proposed approach and potential improvements required for improved performance.

Second, PICO framework based filtration at abstract level is limited to citations having abstracts. However, there are citations (e.g. practice guidelines) where the abstract is not available. Therefore, searching for relevant information should be extended to full text rather than at abstract level.

Third, our approach implements population, intervention, comparison and disease for concept-based screening algorithm. Outcome is not included because we believe that the metrics for primary and secondary outcomes are fairly standard for a given population, intervention and comparison. However, it is unclear whether the performance of our system could improve when outcome is used.

Finally, using UpToDate as a gold standard has a major limitation that all relevant and potentially useful citations from the primary literature are not always available in UpToDate. Therefore, citations retrieved by the system might include potentially useful information that is not available in UpToDate.

**CONCLUSIONS**

The citation retrieval system we have presented for evidence-based clinical knowledge summarization integrates a pipeline of NLP techniques for query expansion, concept-based screening, and relevancy measure with concept-based vector similarity. We showed that the accuracy of citation retrieval is significantly improved by using concept-based vector similarity after extracting population, intervention or comparison, and disease concepts. Furthermore, evaluation of the system on Top K citations achieved an overall F-score of 69.9% on heart failure and 75.1% on atrial fibrillation showing the importance of users' judgment about the number of citations to be used. Overall, the hybrid combination of concept-based screening algorithm for screening citations and for relevancy measure using concept-based vector similarity outperformed approaches using only PICO framework, term-based vector similarity, or query expansion. Annotation of apparent false positives by domain experts shows that few of these citations actually have new knowledge that is not available in online clinical resources such as UpToDate.

**Declaration of interests**

KR, AJS, RPG, MRK, and SRJ declare no competing interests.

**Acknowledgments**

This work was made possible by funding from the National Library of Medicine grant: R00LM011389.